\pdfoutput=1
\documentclass[11pt]{article}

\usepackage[final]{acl}

\usepackage{times}
\usepackage{latexsym}

\usepackage[T1]{fontenc}
\usepackage[utf8]{inputenc}

\usepackage{microtype}

\usepackage{inconsolata}

\usepackage{graphicx}

\usepackage{amsmath,amssymb,amsfonts}%
\usepackage{amsthm}%
\usepackage{mathrsfs}%
\usepackage{subfig}
\usepackage{multirow}

\title{Mapping Patient Trajectories: Understanding and Visualizing Sepsis Prognostic Pathways from Patients Clinical Narratives}
\author{Sudeshna Jana \\ TCS Research, India \\sudeshna.jana@tcs.com 
        \And Tirthankar Dasgupta \\ TCS Research, India \\dasgupta.tirthankar@tcs.com
        \And Lipika Dey\\ Ashoka University, India \\lipika.dey@ashoka.edu.in
      }
\begin{document}
\maketitle
\begin{abstract}
In recent years, healthcare professionals are increasingly emphasizing on personalized and evidence-based patient care through the exploration of prognostic pathways. To study this, structured clinical variables from Electronic Health Records (EHRs) data have traditionally been employed by many researchers. Presently, Natural Language Processing models have received great attention in clinical research which expanded the possibilities of using clinical narratives. In this paper, we propose a systematic methodology for developing sepsis prognostic pathways derived from clinical notes, focusing on diverse patient subgroups identified by exploring comorbidities associated with sepsis and generating explanations of these subgroups using SHAP. The extracted prognostic pathways of these subgroups provide valuable insights into the dynamic trajectories of sepsis severity over time. Visualizing these pathways sheds light on the likelihood and direction of disease progression across various contexts and reveals patterns and pivotal factors or biomarkers influencing the transition between sepsis stages, whether toward deterioration or improvement. This empowers healthcare providers to implement more personalized and effective healthcare strategies for individual patients.
\end{abstract}

\section{Introduction}
In healthcare, there is an increasing trend to shift towards from doctor-centered treatment to patient-centered treatment approaches, where the intent is to design individualized care for patients based on their health conditions, demography, personal history, and preferences \citep{johnson2021precision, wang2021toward, esfahani2020moving}. This approach promises better outcomes for all since any two patients are not exactly similar. Even a simple disease can be heterogeneous in its clinical presentation in terms of multi-morbidity, severity, as well as response to treatments \cite{alexander2021identifying, battaglia2020introducing}. By implementing personalized care plans, healthcare costs can be reduced by eliminating unnecessary medical examinations and tailoring treatment plans accordingly.

Prognostic pathways, derived from the experiences of past patients, play a major role in clinical decision-making in general, and more specifically in enabling personalized treatments. These pathways outline the expected disease progression, stages, influential factors of a particular disease, and potential outcomes for a specific patient or group of patients. It provides valuable guidance and support to healthcare professionals to assess a newly arrived patient's risk of developing complications, how the disease is expected to progress, and the likelihood of certain outcomes. Based on the individual patient's assessed risk, medical practitioners can tailor their treatment approach to meet the patient's unique needs. For instance, for a patient with high risk, medical practitioners may choose a more aggressive treatment approach or monitor the patient more closely. Moreover, by gaining insights into the expected disease trajectory, healthcare professionals can allocate critical healthcare resources such as intensive care units (ICUs), operating rooms (OTs), mechanical ventilators, etc. more efficiently. These prognostic pathways empower healthcare professionals to deliver precision-based, patient-centered care, ultimately enhancing the overall quality of healthcare services.
%First, modeling progression of diseases is challenging because it is often manifested by multiple symptoms over time, which makes it difficult to summarize progression patterns.

\begin{figure}[!h]
  \centering
  \includegraphics[width=\linewidth]{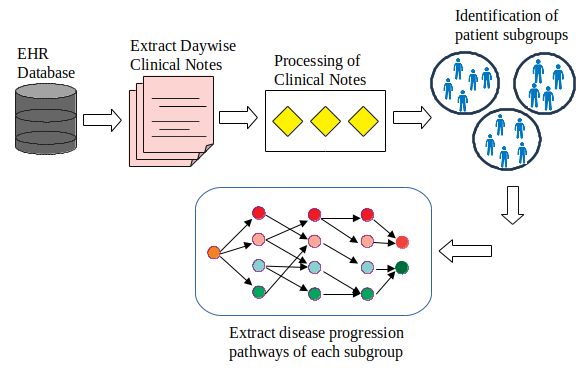}
  \caption{Overview of prognostic pathway development process.}
  \label{overview_of_work}
\end{figure}

In recent years, clinical researchers have increasingly employed diverse disease progression models to analyze and delineate the trajectory of disease development based on longitudinal health records of patients. \citealp{seoane2014pathway} proposed a pathway-based data integration framework for predicting breast cancer progression. Subsequently, \citealp{zhang2015paving} introduced a practice-based clinical pathway development process along with a data-driven methodology to extract common clinical pathways for chronic kidney disease. Also, \citealp{aisen2017path} explored the concept of a disease continuum, examining Alzheimer's disease across pathophysiological, biomarker, and clinical perspectives. \citealp{kwon2020dpvis} developed DPVis, a visual analytics system that integrates Hidden Markov models with interactive visualizations, to explore disease progression patterns from health records. Also \citealp{arias2020mapping} detailed the application of process mining techniques as a valuable tool for evaluating and understanding patients' journeys. In another work, \citealp{zhou2020risk} investigated disease progression in a cohort of 2019-nCoV patients and analyzed associated risk factors. \citealp{vesga2021chronic} conducted a study in 2021 examining the variability in CKD progression and estimating the probability of transition between CKD stages over time. Recently, \citealp{nenova2022chronic} proposed an intelligent case-based reasoning (iCBR) approach for predicting kidney disease progression. Moreover, \citealp{nagamine2022data} introduced a data-driven approach, aiming to generate real-world characteristics and progression patterns of heart failure. These collective efforts showcase the evolving landscape of research methodologies aimed at comprehensively understanding and predicting the progression of various diseases.

The majority of the aforementioned studies have concentrated on the analysis of structured electronic health records (EHR) encompassing numerical and categorical data values derived from vital signs, lab test results, and medication prescriptions. However, clinical notes, intricately linked to patients' EHRs, encapsulate valuable information concerning patients' conditions, symptoms, diagnoses, treatments, chronic and historical ailments, drug prescriptions, adverse effects on patients, etc. Consequently, analyzing such textual data offers the opportunity to gain a deeper understanding of the patient's health condition as well as the physician's rationale behind a chosen treatment path. In this study, we have focused on a cohort of sepsis patients sourced from the publicly available Medical Information Mart for Intensive Care (MIMIC-III v1.4) database \cite{johnson2016mimic}. The main objective of our study is to comprehend various sepsis progression pathways and assess risks for diverse patient subgroups based on real-world clinical practices. In our work, we utilize a collection of time-stamped clinical notes, including radiology and ECG reports, along with nursing notes of patients. The proposed systematic methodology is depicted in Figure \ref{overview_of_work}. This process involves the representation of unstructured clinical notes using the biomedical thesaurus, the identification of distinct patient subgroups, and the extraction of disease progression pathways accompanied by a comprehensive risk analysis.

The subsequent sections of this article are structured as follows. In the next section, we present the detailed methodology for data representation, patient subgroup identification, and prognostic pathway extraction. Following that, we present a concise overview of our study dataset and the outcomes of our experiments. Finally, we give a comprehensive discussion of our analytical findings, draw a conclusion from our analysis. 

\section{Proposed Methodology}
% \subsection{Study Population}
% The study is performed on approximately 1,500 sepsis patients, sourced from the MIMIC-III v1.4 database \cite{johnson2016mimic}. This extensive database encompasses the medical records of over forty thousand patients diagnosed with various diseases between 2001 and 2012 at the Beth Israel Deaconess Medical Center (BIDMC). It integrates both structured and unstructured clinical events documented during hospital admissions. Notably, the database adheres to rigorous anonymization protocols, ensuring meticulous protection of patient privacy. Furthermore, to enhance privacy measures, specific dates and times of events have been intentionally obscured. The database holds pre-existing Institutional Review Board (IRB) approval, and researchers gain access to the data upon successful completion of the 'Data or Specimens Only Research' training course provided by the Collaborative Institutional Training Initiative (CITI).

% We have extracted key demographic characteristics from patients, including age, gender, and length of stay alongside comprehensive clinical notes, such as nursing notes, radiology reports, ECG reports, and discharge summaries recorded throughout the entire hospitalization period.

In this section, we present a detailed description of our proposed systematic methodology for generating prognostic pathways from patients' day-to-day textual clinical reports such as nursing notes, radiology reports, and ECG reports. In previous works \cite{jana2022using, jana2022predicting}, authors utilized several transformer-based representations such as BERT, ClinicalBioBERT, and BlueBERT embeddings of these notes in various predictive models. While these embeddings effectively captured linguistic nuances like distinguishing between severe and mild pain, they sometimes struggled to discern similarities or differences between two notes based solely on medical terms. Therefore, before getting into the stratification work, where note similarity is crucial, we introduced an additional processing layer. Each clinical note underwent initial processing through biomedical dictionaries to standardize terms. The details of the processing pipeline using the biomedical dictionaries are presented below. 

\subsection{Transformation of Unstructured Clinical Notes into Structured Representations}\label{Unstructured to Structured}
%\subsection{Data Transformation}
\begin{figure}[!h]
  \centering
  \includegraphics[width=\linewidth]{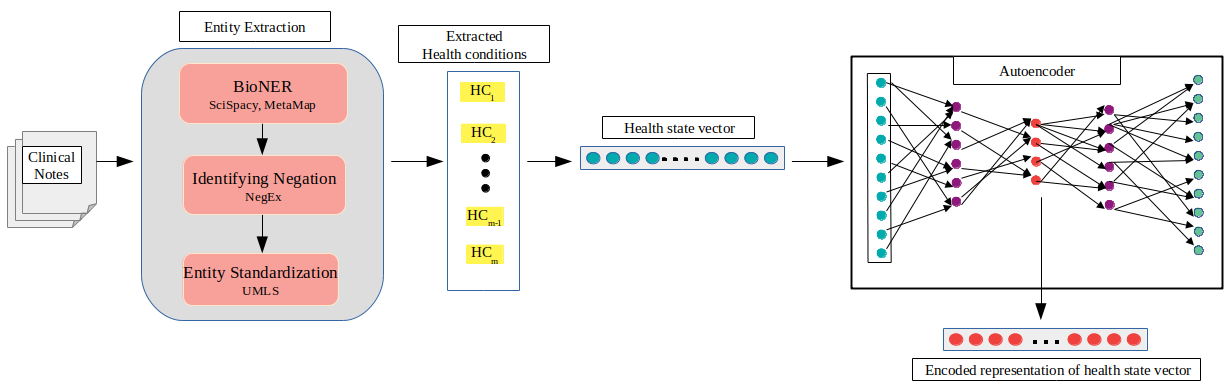}
  \caption{Transformation of unstructured clinical notes to structured representation.}
  \label{structured_representation}
\end{figure}

Clinical notes, specifically nursing documentation, display significant diversity in both style and content. Some healthcare professionals document solely the symptoms present on a given day, while others meticulously record the absence of common symptoms, adverse reactions, the psychological state of patients, appetite changes, and more. The utilization of non-standard terminology and abbreviations is also frequently observed in these notes. To address this diversity, we have introduced an additional processing layer, wherein each clinical note undergoes initial processing through the Biomedical dictionaries to derive a more structured representation of the patients' health conditions, as illustrated in Figure \ref{structured_representation}. The details of the processing pipeline using the biomedical dictionaries are presented below.

\subsubsection{Entity Extraction:}
We employed two BioNER tools, ScispaCy \cite{neumann2019scispacy} and Metamap \cite{aronson2006metamap}, for the extraction of patients' health conditions from clinical notes. The pre-trained scispaCy model, specifically \textit{en\_ner\_bc5cdr\_md}, was utilized for recognizing ``disease'' names mentioned in clinical notes. Simultaneously, through the use of Metamap, we identified eight medical entities, including ``Sign or Symptom'', ``Disease or Syndrome'', ``Acquired Abnormality'', ``Anatomical Abnormality'', ``Congenital Abnormality'', ``Injury or Poisoning'', ``Mental Process'', and ``Mental or Behavioral Dysfunction'' within these notes.

We have also extracted the final recovery status of patients from the discharge summaries. After analyzing the descriptions, we categorized patients into two major states at the time of discharge: `Decease', and `Discharge'.

\subsubsection{Detecting Negations:}
Subsequently, the Negex algorithm \cite{chapman2001simple}, designed to identify negative modifiers such as ``no'', ``not'', etc., is employed to detect negative mentions of entities within the text. The initial list was expanded to encompass commonly occurring negation concepts like `deny'', ``refuse'', ``absent'', ``decline'', etc., frequently encountered in clinical notes. For instance, in a sentence like ``The patient has shortness of breath but denies any chest pain'', the two symptoms identified would be ``shortness of breath'' and ``neg chest pain''. These negative symptoms play a crucial role in providing a comprehensive understanding of individual patients.

\subsubsection{Clinical Entity Normalization:}
Clinical notes often encompass diverse non-standard terminology, abbreviations, various formats, and coding systems to represent clinical concepts. For instance, a single medical condition like ``Hemorrhage'' may be referred to as ``Bleeding'', ``Blood loss'' or ``oozing of blood'' by different healthcare professionals. To address this variability, we have standardized all extracted entities using the UMLS Metathesaurus \cite{schuyler1993umls}, which includes a comprehensive list of such scenarios and assigns a ``Concept Unique Identifier (CUI)'' to each. However, we observed that certain entities did not yield an exact match with any UMLS concept. To resolve this, an approximate string-matching algorithm was employed, identifying the closest UMLS concept based on the Levenshtein distance measure \cite{yujian2007normalized} for entities without an exact match. In cases where entities couldn't be mapped to any UMLS concept, unique identifiers were created to ensure no health condition was overlooked. To prevent any ambiguity, we explicitly refer to these unique identifiers as CUIs.

Now, every clinical note can be effectively represented by the presence or absence of CUIs. Let the comprehensive list or vocabulary of CUIs, encompassing descriptions of diseases and symptoms relevant to a specific study, be denoted as $V$. Consequently, a patient's condition at a particular point in time can also be expressed in terms of these CUIs.

\subsubsection{Handling Missing Data:}
In our analysis of EHR, we identified a common challenge related to the absence of documented medical records on certain hospital days, leading to a lack of insights into the patient's medical actions during those periods. Additionally, incomplete medical records in clinical notes pose another issue. For instance, information about a specific disease (e.g., urinary tract infection) may be mentioned in $Day_{n-1}$ notes and in $Day_{n+1}$ but was not mentioned in $Day_{n}$ notes, creating uncertainty about the presence of that disease in the prognostic pathway. To overcome these issues, we have defined the following rules to gain insights into missing days and ensure a continuous understanding of the patient's condition:

\begin{enumerate}
  \item If a disease or symptom $d$ is present in $Day_{n-1}$ and $Day_{n+1}$, we consider it to be present in $Day_{n}$ as well.
  \item If a disease or symptom $d$ is noted as negative in $Day_{n-1}$ and $Day_{n+1}$, we assume it is also negative in $Day_{n}$.
  \item If a disease or symptom $d$ is present in $Day_{n-1}$ and negative in $Day_{n+1}$, we assume it is positive in $Day_{n}$.
  \item If a disease or symptom $d$ is noted as negative in $Day_{n-1}$ and never occurred in the future, we consider it to be negative in all future days.
\end{enumerate}
By applying these rules, we aim to alleviate the impact of missing or incomplete data, providing a more comprehensive understanding of the patient's medical history and progression.

\subsubsection{Vector Representation:}
Afterward, we have segmented the patient's hospitalization duration into distinct stages. We defined the initial stage, or stage 1, as encompassing the diseases or symptoms observed on the first two days. The day of discharge marked as the final stage or discharge stage. The intervening days between the initial and discharge stages were further divided into three-day windows, forming subsequent stages. 
%consequently, the patient’s health condition during any specific stage can also be represented by consolidating CUIs observed across these days.

Given a patient $p$, the health condition at stage $t$ is defined by a vector $H_p(t) = <d_i>$ , $i = 1, 2,..., |V|$ 
, where $d_i \in V$ and
\[
d_i =
\begin{cases}
1 & \text{if $d_i$ present in stage $t$ for patient $p$} \\
-1 & \text{if $d_i$ negative in stage $t$ for patient $p$}\\
0 & \text{if $d_i$ not mentioned in stage $t$ for $p$} 
\end{cases}
\]

As the number of unique diseases or symptoms obtained from any patient dataset is very high and individuals may not manifest all symptoms or diseases, the resulting vectors are characterized by high dimensionality and sparsity. To overcome this issue, we have employed an autoencoder-based transformation \cite{wang2016auto} to obtain a dense representation in a lower-dimensional space. In an autoencoder (AE) model, the ``encoder'' network creates a compressed representation of the input data by capturing the essential characteristics and underlying patterns, while the ``decoder'' network learns to reconstruct the original input data from the compressed representation while minimizing the loss of information. The resulting compressed representations serve as the vector representation of the patient's health conditions for our further work.

\subsection{Identification of Patient Subgroups Based on Initial Health Conditions}
We expect significant diversity among patients with varying comorbidities. Therefore, before doing the risk assessment, we aim to categorize patients into subgroups based on their health conditions at the initial stage. This helps in understanding for which comorbidities patients will be in high-risk or low-risk in the future.
%This approach will enable us to extract tailored prognostic pathways for distinct patient types, ensuring a more personalized and targeted analysis.
In our study, we have used the k-means clustering algorithm \cite{ja1979k}, utilizing the Euclidean distance as the metric to assess similarities among patients. For a given value of k, a set of k cluster centers is randomly selected, and each data point is assigned to the cluster by iteratively minimizing the within-cluster distance. To determine the optimal value of k, we have utilized the silhouette coefficient \cite{kodinariya2013review}. This coefficient measures how similar each point is to others within the same cluster compared to points in other clusters. The average silhouette coefficient, computed across all points, offers a metric for assessing the cohesiveness of each cluster as well as their separation or distinctiveness from one another.
%Starting with 2, the value of k is iteratively increased as long as the silhouette score also increases with it. The ideal value of k is the one that yields the highest average silhouette score, beyond which the score starts to decrease steadily.

To generate human-interpretable explanations for the clusters, we have proposed leveraging Shapley values \cite{merrick2020explanation}, which quantify the contribution of each feature for each individual towards the final outcome while preserving the sum of all contributions. Our objective was to provide explanations in terms of diseases or symptoms, encompassing the predominant symptoms within a cluster and highlighting the differentiating aspects between clusters. We utilized a CUI-based representation for this purpose. By treating cluster labels as target outcomes, we trained a Random Forest classifier to predict these labels using the CUI vector-based representation of patients. The resulting model was analyzed using the SHAP TreeExplainer to gain insights into the decision-making process. This method not only reveals the contribution of each symptom to a specific label but also provides SHAP values for each patient, facilitating the interpretation of why a patient has been assigned to a particular cluster. Moreover, it also helps in the interpretation of misclassifications by the model, if any.

\subsection{Extracting Prognostic Pathways for Patient Subgroups}
In our study, we present a comprehensive explanation of the process of extracting progression networks for sepsis patients, which depict the transition states for each stage across various patient subgroups in sepsis, recognized as a form of prognostic pathway. To extract prognostic pathways, our initial step involves the identification and categorization of sepsis severity for each patient in each stage according to the Sepsis-3 definition \cite{singer2016third}. The Sepsis-3 criteria, introduced by the Third International Consensus Definitions for Sepsis and Septic Shock (Sepsis-3) in 2016, provides a clinical framework for assessing sepsis severity and classifying patients into four distinct states: Systemic Inflammatory Response Syndrome (SIRS), Sepsis, Severe Sepsis, and Septic Shock. 

For the computation of sepsis severity at each stage, structured features such as temperature, heart rate, respiratory rate, and white blood cell (WBC) count were systematically extracted from the `CHARTEVENTS.csv' file within the MIMIC database.
%The most critical values for each of these parameters were considered in determining the severity at each stage. 
Additionally, complementary information related to infection, organ dysfunction, hypotension, intravenous (IV) fluid resuscitation, and other relevant features is derived from our previously collected data obtained from clinical notes. This integration of structured and unstructured data enhances the comprehensiveness of our approach and provides a more nuanced understanding of sepsis severity across different stages. To quantitatively represent the severity of each sepsis state, we have assigned a severity score to each: SIRS: 1, Sepsis: 2, Severe Sepsis: 3, and Septic Shock: 4. This scoring system enhances the interpretability of our findings and facilitating a clearer communication of the severity levels associated with each sepsis state.

When a patient transitions from one stage to the next, we have defined the potential outcomes or states based on the progression of sepsis severity as follows:
\begin{itemize}
    \item \textbf{Discharge}: when the patient is discharged in the next stage.
    \item \textbf{Improve}: when the severity score decreases compared to the previous stage, indicating a positive response.
    \item \textbf{Persistent}: if the severity score remains unchanged from the previous stage.
    \item \textbf{Deteriorate}: when the severity score increases compared to the previous stage, signifying a worsening condition.
    \item \textbf{Decease}: if the patient is expired during the next stage.
    \item \textbf{Unknown}: if the sepsis state is unknown in the next stage, due to missing information in the database.
\end{itemize}

Afterward, we have analyzed the outcomes or states during the transition for each stage across different patient subgroups. In the progression networks of sepsis severity for each patient subgroup, each stage, except final stage consists five nodes represent distinct states such as discharge, improvement, persistent, deterioration and decease. In the final stage, only two nodes, discharge and decease, remain. To simplify our analysis, we excluded the `Unknown' state. The edges, denoted as $e_{s_is_j}$, represent the probability of transitioning to state $s_j$ in the next stage based on the state of preceding stage $s_i$, expressed as $e_{s_is_j} = P(X_t = s_j|X_{t-1} = s_i)$. These networks provide a visual representation of how sepsis severity changes through different stages of the disease, offering insights into the potential trajectories and outcomes for patients within specific subgroups. 

%We have also extracted the final recovery status of patients from the discharge summaries. After analyzing the descriptions, we categorized patients into two major states at the time of discharge: (a) Decease, and (b) Discharge. In the subsequent section we provide a comprehensive explanation of our prognostic pathways extraction process from a cohort of sepsis patients.

\section{Results and Discussions}

\subsection{Study Population}
The study is performed on a cohort of `Sepsis' patients, sourced from the MIMIC-III v1.4 database \cite{johnson2016mimic}. This extensive database encompasses the medical records of over forty thousand patients diagnosed with various diseases between 2001 and 2012 at the Beth Israel Deaconess Medical Center (BIDMC). It integrates both structured and unstructured clinical events documented during hospital admissions. Notably, the database adheres to rigorous anonymization protocols, ensuring meticulous protection of patient privacy. The database holds pre-existing Institutional Review Board (IRB) approval, and researchers gain access to the data upon successful completion of the `Data or Specimens Only Research' training course provided by the Collaborative Institutional Training Initiative (CITI). In our study, we specifically focused on 1593 sepsis patients, excluding the rest due to very short lengths of stay (i.e., less than 1 day) or insufficient information for most of the hospital days. Within this selected cohort, $54\%$ were male, and $46\%$ were female. Only $0.1\%$ were under the age of 18, $7\%$ were between 18-40 years old, $23\%$ were between 41-60 years old, $43\%$ were between 61-80 years old, and $25\%$ were over 80 years old. The average length of stay for this cohort was 11 days.

\subsection{Subpopulations within Sepsis Patient Cohort}

\begin{figure}[!h]
  \centering
  \includegraphics[width=\linewidth]{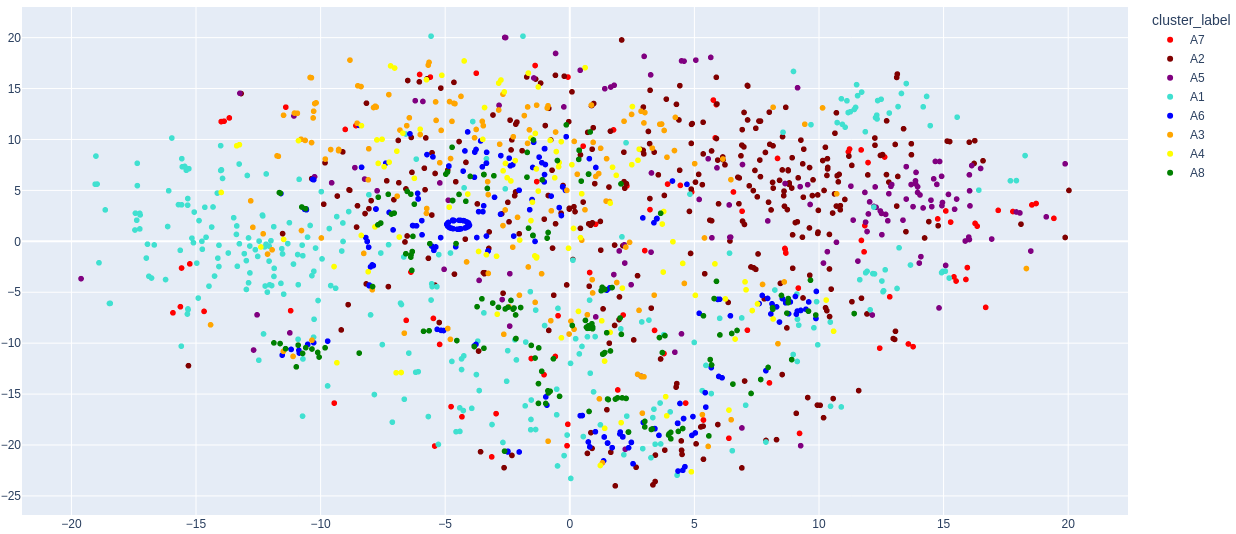}
  \caption{Distribution of 8 clusters using 2D t-SNE visualization.}
  \label{8_clusters}
\end{figure}

From the selected cohort, a total of 19,543 clinical notes, encompassing nursing notes, ECG reports, and radiology reports were extracted. Following the pre-processing steps outlined in Section \ref{Unstructured to Structured}, we compiled a comprehensive list of 3500 unique diseases or symptoms. Subsequently, we have segmented each patient's hospitalization days into stages, as previously discussed, resulting in a maximum of 11 stages. After this segmentation, we generated 500-dimensional auto-encoded vectors for the health conditions of the initial stage and obtained 8 distinct patient subgroups, as depicted in Figure \ref{8_clusters}. In Table \ref{8 patient subgroups}, we present summary statistics and highlight key diseases or symptoms obtained from the SHAP Explainer across these identified subgroups.

%From this dataset, a total of 19,543 clinical notes, encompassing nursing notes, ECG reports, and radiology reports were extracted. Following the pre-processing steps outlined in Section \ref{Unstructured to Structured}, we compiled a comprehensive list of 3500 unique diseases or symptoms. Subsequently, we have segmented each patient's hospitalization days into stages, as previously discussed, resulting in a maximum of 11 stages. After this segmentation, we generated 500-dimensional auto-encoded vectors for the health conditions of the initial stage.

%Then we have identified 8 distinct patient subgroups, as depicted in Figure \ref{8_clusters} based on the auto-encoded health vectors in the initial stage. In Table \ref{8 patient subgroups}, we present summary statistics and highlight key diseases or symptoms obtained from the SHAP Explainer across these identified subgroups.

\begin{table*}[!t]
\centering
\footnotesize
%\resizebox{.95\columnwidth}{!}{
\renewcommand{\arraystretch}{1.5}
\begin{tabular}{|p{.5in}|p{.5in}|p{5in}|} 
\hline
Subgroup & $\#$Patients & prominent diseases or symptoms\\
\hline 
A1 & 298 & sepsis with hypotension, acidosis, diabetes, respiratory distress, pain, tachycardia\\ 
\hline
A2 & 339 & sepsis with loose stool, hypotension absence of acidosis, pain, fever\\
\hline
A3 & 155 & sepsis with dyspnea, pain, hypotension, airway disease absence of diabetes, acidosis\\
\hline
A4 & 105 & sepsis with hypotension, skin infection, pain, urinary tract infection, kidney diseases\\
\hline
A5 & 128 & sepsis with basilar rales, dyspnea, hypotension, edema, premature ventricular contraction (PVC), urinary tract infection (UTI), heart disease\\
\hline
A6 & 284 & sepsis with tachycardia, atrial fibrillation, atrial premature complexes \\
\hline
A7 & 91 & sepsis with premature ventricular contraction (PVC), hypotension, thick sputum, loose stool, diabetes, erythema, basilar rales, atrial fibrillation\\
\hline
A8 & 193 & sepsis with myocardial infarction, bundle-branch block, ventricular hypertrophy, anterior fascicular block\\

\hline

\end{tabular}
\caption{Summary of 8 patient subgroups based on initial health conditions obtained from the SHAP Explainer.}
\label{8 patient subgroups}
\end{table*}

%\subsubsection{Identification of Sepsis Severity}

\subsection{Sepsis Prognostic Pathways}

\begin{figure}[!h]
  \centering
  \includegraphics[width=\linewidth]{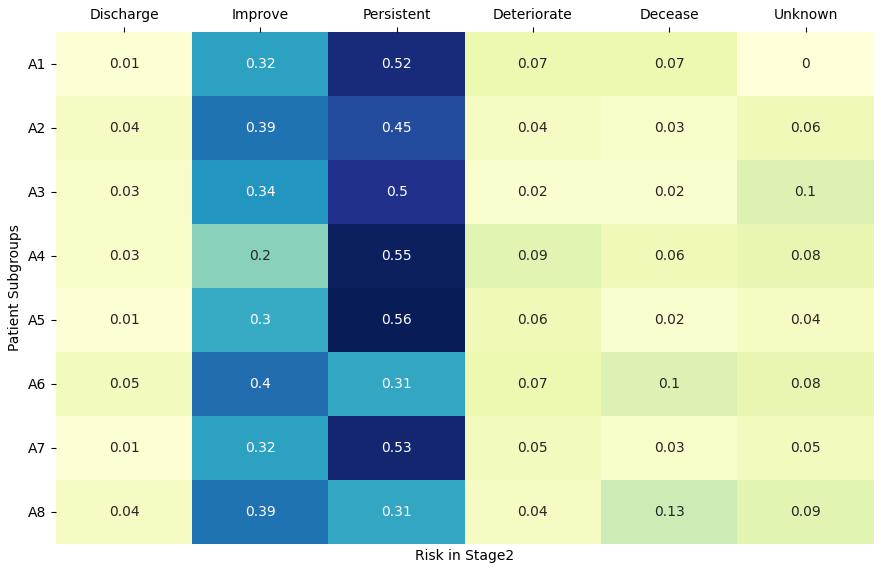}
  \caption{Heatmap displaying transition probabilities of different states after 2 days of admission for each patient subgroup.}
  \label{Stage2}
\end{figure}

In this section, we have analyzed the associated risks in sepsis progression in terms of outcomes or states during the transition from one stage to the next across eight distinct patient subgroups. Figure \ref{Stage2} illustrates the transition probabilities of different states after 2 days of admission for each of these patient subgroups, obtained from our dataset.

% \begin{figure*}
% \centering
% \subfloat[Patient subgroup A1]{\includegraphics[width = 2.5in]{figures/stage2-stage3/A1_1.png}} 
% \subfloat[Patient subgroup A2]{\includegraphics[width = 2.5in]{figures/stage2-stage3/A2_1.png}}\\
% \subfloat[Patient subgroup A3]{\includegraphics[width = 2.5in]{figures/stage2-stage3/A3_1.png}}
% \subfloat[Patient subgroup A4]{\includegraphics[width = 2.5in]{figures/stage2-stage3/A4_1.png}}\\
% \subfloat[Patient subgroup A5]{\includegraphics[width = 2.5in]{figures/stage2-stage3/A5_1.png}}
% \subfloat[Patient subgroup A6]{\includegraphics[width = 2.5in]{figures/stage2-stage3/A6_1.png}}\\
% \subfloat[Patient subgroup A7]{\includegraphics[width = 2.5in]{figures/stage2-stage3/A7_1.png}}
% \subfloat[Patient subgroup A8]{\includegraphics[width = 2.5in]{figures/stage2-stage3/A8_1.png}}

% \caption{Heatmap visualization and probability of risks in stage 3 based on the states of each subgroup in stage 2. In each sub-figure, the Y-axis represents the states in stage 2, and the X-axis represents the states in stage 3. }
% \label{stage3}
% \end{figure*}

% \begin{figure*}[!h]
%   \centering
%   \includegraphics[width=\linewidth]{figures/patient_health_progression_network_A3.png}
%   \caption{Progression network of sepsis severity for patient subgroup A3. Edge color indicates transition probabilities: black for probabilities $\geq$ 0.5, red for probabilities between 0.3 to 0.5, violet for probabilities between 0.1 to 0.3, and turquoise for probabilities $<$ 0.1.}
%   \label{Sepsis_state_progression_A3}
% \end{figure*}

\begin{figure*}[!h]
  \centering
  \includegraphics[width=\linewidth]{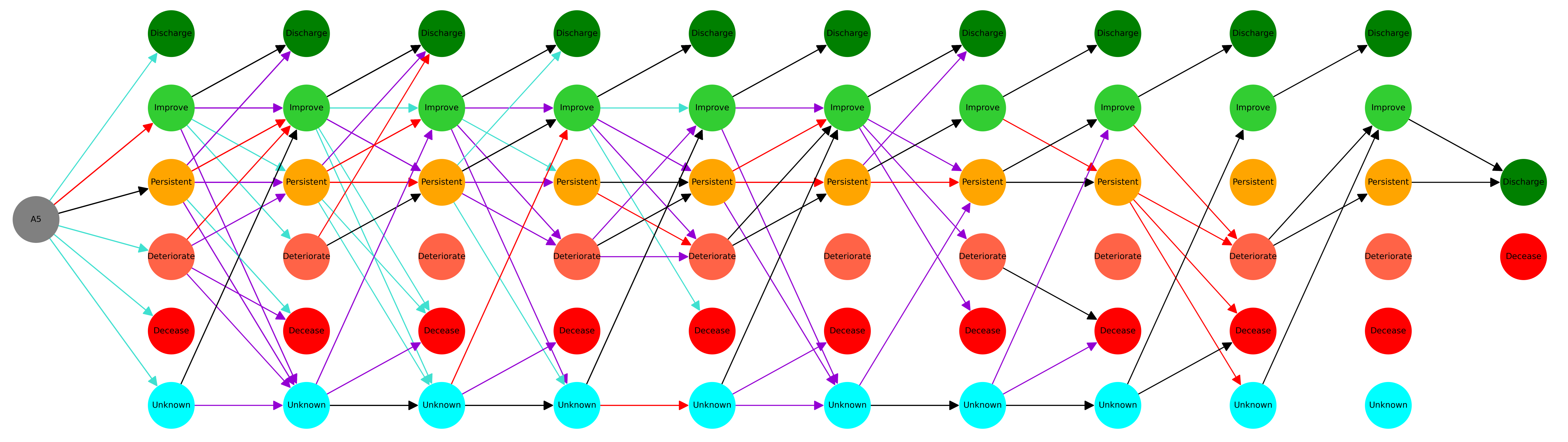}
  \caption{Progression network of sepsis severity for patient subgroup A5. Edge color indicates transition probabilities: black for probabilities $\geq$ 0.5, red for probabilities between 0.3 to 0.5, violet for probabilities between 0.1 to 0.3, and turquoise for probabilities $<$ 0.1.}
  \label{Sepsis_state_progression_A5}
\end{figure*}

% \begin{figure*}[!h]
%  \centering
%  \includegraphics[width=\linewidth]{figures/A3.png}
%  \caption{Disease progression in subgroup A3 and its impact on Sepsis Severity. Edge labels provide insight into the two most effectively treated conditions (highlighted in green) and the top two newly emerging diseases (highlighted in red) for each transition.}
%  \label{disease_progression_A3}
% \end{figure*}

\begin{figure*}[!h]
 \centering
 \includegraphics[width=\linewidth]{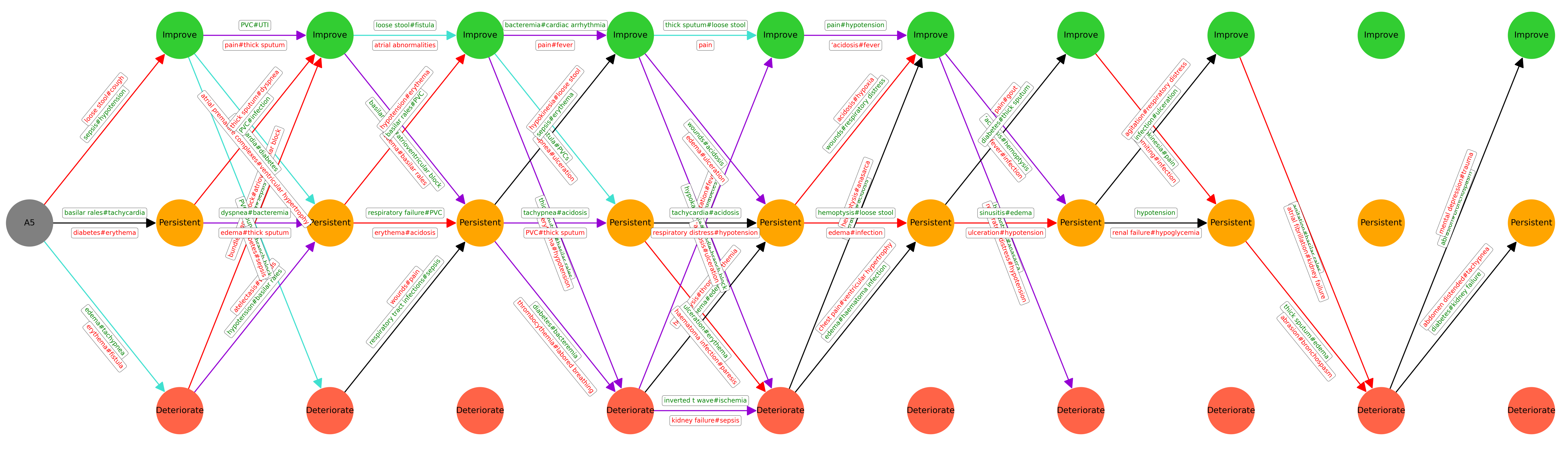}
 \caption{Disease progression in subgroup A5 and its impact on Sepsis Severity. Edge labels provide insight into the two most effectively treated conditions (highlighted in green) and the top two newly emerging diseases (highlighted in red) at end of the transition.}
 \label{disease_progression_A5}
\end{figure*}

\begin{table*}[!t]
\centering
\footnotesize
%\resizebox{.95\columnwidth}{!}{
\renewcommand{\arraystretch}{1.5}

\begin{tabular}{|p{1.5in}|p{3in}|p{.5in}|} 
\hline
Model & Features used & Accuracy\\
\hline
\multirow{2}{1in}{Predict stage1 subgroup using Random Forest} 
& BlueBERT representation of notes & 85\% \\
& \textbf{our representation of notes} & \textbf{89\%} \\ 
\hline
\multirow{4}{1in}{Predict state in stage2 using NN classifier}
& BlueBERT representation of notes & 43\% \\
& our representation of notes & 70\% \\ 
& BlueBERT representation of notes + stage1 subgroup label & 48\% \\
& \textbf{our representation of notes + stage1 subgroup label} & \textbf{75\%} \\
\hline
\end{tabular}
\caption{Performance analysis across different representations of clinical notes for stage1 subgroup prediction and stage2 state prediction.}
\label{Performance_of_prediction_model}
\end{table*}

In this figure, we have observed that, in subgroups A2, A6, and A8, although a relatively small number of patients were discharged in Stage 2, approximately 40\% of patients exhibited an improvement in sepsis severity during this stage. Notably, in subgroups A6 and A8, 10\% and 13\% of patients, respectively, experienced unfortunate outcomes and deceased during the second stage. However, for the majority of patients in each group, the severity of sepsis remained consistent.
%Figure \ref{stage3} illustrates the transition probabilities of risks for each subgroup from Stage 2 to Stage 3.
Similarly, we have analyzed the penitential outcomes or states for each subgroup during the transitions between other stages also. We have noticed that, patients across subgroups A5, A2, and A8, whose sepsis severity improved in Stage 2, exhibited the highest discharge rates at 54\%, 49\% and 48\%, respectively, indicating a complete recovery in Stage 3. In contrast, patients in Subgroup A1, with improved sepsis severity in Stage 2, exhibited the lowest discharge rate at 11\%, with a majority experiencing unchanged sepsis severity. 
%Furthermore, we observed that 40\% of patients in subgroup A7, 25\% in A5, 24\% in A1, 19\% in A4, 15\% in A2, and 12\% in A8 experienced worsening sepsis severity in Stage 2 and unfortunately deceased in Stage 3. 
Furthermore, we observed that 40\% of patients in subgroup A7 experienced worsening sepsis severity in Stage 2 and unfortunately deceased in Stage 3. Remarkably, no patients in A3, A4, and A6 experienced deterioration-related mortality in Stage 3. 
%Specifically, patients in subgroup A3, whose sepsis severity deteriorated in Stage 2, all demonstrated improvement in Stage 3. 
%However, 15\% of patients in subgroup A1 and 19\% of patients in subgroup A4, despite experiencing improved sepsis severity in Stage 2, unfortunately, deteriorated in Stage 3.
This finding highlights diverse outcomes among distinct subgroups of sepsis patients.

Figures \ref{Sepsis_state_progression_A5} depict the full progression network, which provides a visual representation of how sepsis severity changes through different stages for patient subgroups A5. %In these figures, each stage, from 2 to 11, consists three nodes represent distinct states such as discharge, improvement, persistent, deterioration and decease. In the final stage, only two nodes, discharge and decease, remain. The edges, denoted as $e_{r_ir_j}$, represent the probability of transitioning to risk $r_j$ in the next stage based on the risk of preceding stage $r_i$, expressed as $e_{r_ir_j} = P(X_t = r_j|X_{t-1} = r_i)$. 
The color gradient of the edges reflects transition probabilities, ranging from high to low. 
%These networks provide a visual representation of how sepsis severity changes through different stages, offering insights into the potential trajectories and outcomes for patients within specific subgroups. 
Moreover, we conducted a detailed analysis of each stage transition in sepsis, by specifically examining patients' diseases and symptoms.
In the figure \ref{disease_progression_A5}, we have showed the two most effectively treated conditions and the top two newly emerging diseases for each transition. We excluded transitions leading to the `Discharge' or `Deceased' states, as these represent the end stages.

%Moreover, we conducted a detailed analysis of each stage transition in sepsis, specifically examining patients' diseases and symptoms. Figure \ref{disease_progression_A3} visually illustrates the disease progression for patient subgroups A3, respectively. To simplify our analysis, we excluded transitions leading to the `Discharge' or `Deceased' states.
%, as `Discharge' indicates stable medical conditions post-treatment, while the `Deceased' state signifies severe deterioration
%In this figure \ref{disease_progression_A3}, we have showed the two most effectively treated conditions and the top two newly emerging diseases for each transition. 
%For patients in subgroup A3, where sepsis severity improved in stage 2, 90\% did not have sepsis, 76\% were free from hypotension, 73\% were without fever, and 50\% did not exhibit dyspnea in stage 2. However, 48\% displayed edema, and 40\% exhibited diabetes during the same stage. Notably, some of these patients were free from diabetes in stage 3, and the sepsis stage also improved. 
%Conversely, among patients in subgroup A3 whose sepsis severity worsened during stage 2, 60\% experienced diarrhea, and 50\% had organ dysfunction in this stage. Fortunately, these conditions improved in stage 3. 

We have observed that, in subgroup A5, patients whose sepsis severity improved in stage 2 were entirely free from sepsis, 92\% were without hypotension, and 72\% did not exhibit tachycardia in stage 2.  However, for a subset of these patients, sepsis recurred with diabetes in stage 3, and conditions deteriorated in this stage.
Conversely, within subgroup A5, patients whose sepsis severity worsened during stage 2, although all were free from tachypnea, and 67\% were without edema. However, among them, erythema occurred in 70\% of patients, and 50\% developed fistula during this stage. Significantly, we noted an improvement in symptoms related to heart diseases, including premature ventricular contraction, atrioventricular block, bundle-branch block, etc., from stage 3. 
The insights gained from the analysis of disease progression enable us to identify patterns and pivotal factors that play a crucial role in the transition between sepsis stages, particularly among diverse patient subgroups, emphasizing the need for personalized care strategies based on the specific characteristics of each subgroup.

\subsection{Next State Prediction for New Patients}

Additionally, we have developed a predictive model aimed at forecasting the progression of sepsis and evaluating future risks for a new set of patients admitted with sepsis. To accomplish this, we first employ various machine learning algorithms such as decision trees, random forests, and XGBoost models to predict the initial stage or `stage 1 cluster' that best corresponds to the patient's current health conditions extracted from clinical notes upon admission. We have obtained an accuracy of 89\% using random forest classifier. Subsequently, we predict the next potential outcome or state in stage 2 for these patients, determining whether their sepsis condition will ``Improve'', ``Persist'', or ``Deteriorate''. We have experimented with different machine learning and deep learning classifiers and the performance of the predictive framework is compared across different representations of clinical notes. In Table \ref{Performance_of_prediction_model}, we present the performance these two predictive models using different input representation.
%When utilizing an autoencoded representation of the first day's clinical notes as discussed above, our model achieves best accuracy of 75\%, with stage 1 cluster ids with neural network classifier. This accuracy drops to 70\% when stage 1 cluster information is not considered. Additionally, we have explored transformer-based representations, such as BlueBERT embeddings of the entire clinical notes. Here, we achieve an accuracy of 48\% and 43\% with and without utilizing stage 1 cluster information for patients, respectively. 
This results shows that the our representation of clinical notes discussed in section \ref{Unstructured to Structured} leads to better prediction performance compared to using transformer-based representations, such as BlueBERT embeddings of the raw clinical notes. 
%These results indicate that incorporating structured representations of health conditions leads to better prediction performance compared to using raw clinical notes alone.
Moreover, integrating cluster information into the models consistently enhanced predictive performance across all representation types. Similarly, we can predict the potential states in subsequent stages based on the health conditions and outcome state from the preceding stage, ultimately providing insight into the potential progression pathway for a new patient.

\section{Conclusion}
In summary, the development of practice-based prognostic pathways offers a structured approach to delivering high-quality, cost-effective care while promoting shared decision-making and facilitating continuous improvement in healthcare delivery. In our study, we have demonstrated the effectiveness of deep learning-based representations in capturing the complexity of clinical notes, thereby providing valuable insights into patient cohorts. Additionally, we have generated comprehensive trajectories for each cohort using these representations. 
%These trajectories, when combined with outcomes, aid in patient risk assessment and guide towards low-risk pathways. 
Furthermore, we are exploring the utility of large language models, such as MedLM, for extracting information from clinical notes. 
%We are also in the process of integrating structured data along with unstructured clinical notes. This multimodal approach will enhance the comprehensiveness of our methodology and offer a more nuanced understanding of sepsis severity across various stages. 
In our future work, we also plan to integrate treatment information, such as medications or procedures, into these prognostic pathways. This integration will enable a deeper understanding of complete clinical pathways, allowing us to identify the most effective treatment strategies and assess any potential adverse effects of drugs that may lead to prolonged hospitalization.

% Bibliography entries for the entire Anthology, followed by custom entries
%\bibliography{anthology,custom}
% Custom bibliography entries only
%\bibliography{custom}

\end{document}